\documentclass{article}
\usepackage{CJKutf8}
\usepackage{spconf,amsmath,amssymb,graphicx,multirow,tabularx,booktabs,float,afterpage,balance,hyperref}

\makeatletter
\newcommand{\fixedsubsection}[1]{%
  \par
  \edef\restoresubsectionparskip{\parskip=\the\parskip\relax}%
  \parskip=0pt
  \@startsection{subsection}{2}{\z@}%
    {-3.25ex\@minus -.2ex}{1.5ex}{\normalfont\large\bfseries}{#1}%
  \AddToHookNext{para/begin}{\restoresubsectionparskip}}
\makeatother

\title{Trajectory-Guided Tokenization of Complex CSI for Wi-Fi Sensing}
\name{\shortstack{Ziyi Wang$^{1,2}$, Kenuo Xu$^{3}$, Jichu Jiang$^{3}$, Yumeng Yang$^{2}$,
Zheng Chen$^{3}$, Xiaofei Bai$^{3}$,\\
Muge Chen$^{3}$, Xuyang Chen$^{2,4}$, Jinglei He$^{4}$, Jannik Hammel Nielsen$^{2}$\textsuperscript{*},
Stefan Schmid$^{1}$\sthanks{Corresponding authors: Jannik Hammel Nielsen and Stefan Schmid.}}}
\address{
$^{1}$Technische Universit{\"a}t Berlin, Berlin, Germany\quad
$^{2}$Huawei Heisenberg Research Center (Munich)\\
$^{3}$Huawei Technologies Co., Ltd., Shanghai, China\quad
$^{4}$Technical University of Munich, Munich, Germany
}
\begin{document}
%
\begingroup
\setlength{\tabcolsep}{0pt}
\maketitle
\endgroup
\begin{abstract}
Wi-Fi channel state information (CSI) enables contactless presence detection
and gesture recognition. Its high-dimensional complex-valued time series
require input representations that preserve informative temporal variations
during compression. We propose Trajectory-Guided Tokenization (TGT), which
combines complex trajectory decomposition with asymmetric attention to
construct compact continuous tokens. For each antenna link and subcarrier,
an orthonormal Helmert transform decomposes short, ordered temporal patches
into local-center and centered-trajectory coordinates. Keys are learned from
the centered-trajectory coordinates, while values retain both components.
Learnable queries aggregate
subcarriers into frequency slots, which are fused into temporal tokens.
Trained jointly from scratch, TGT with TokenMLP achieves the highest mean
accuracy of 92.83\% among all evaluated frontend--backend combinations on the
self-collected dataset. Experiments on EHUNAM and Widar
further support the applicability of TGT to cross-domain
presence detection and gesture recognition.
\end{abstract}
%
\begin{figure*}[!t]
\centering
\begingroup
\makeatletter
\let\Gread@transgrouptrue\Gread@transgroupfalse
\makeatother
\includegraphics[page=1,width=\textwidth]{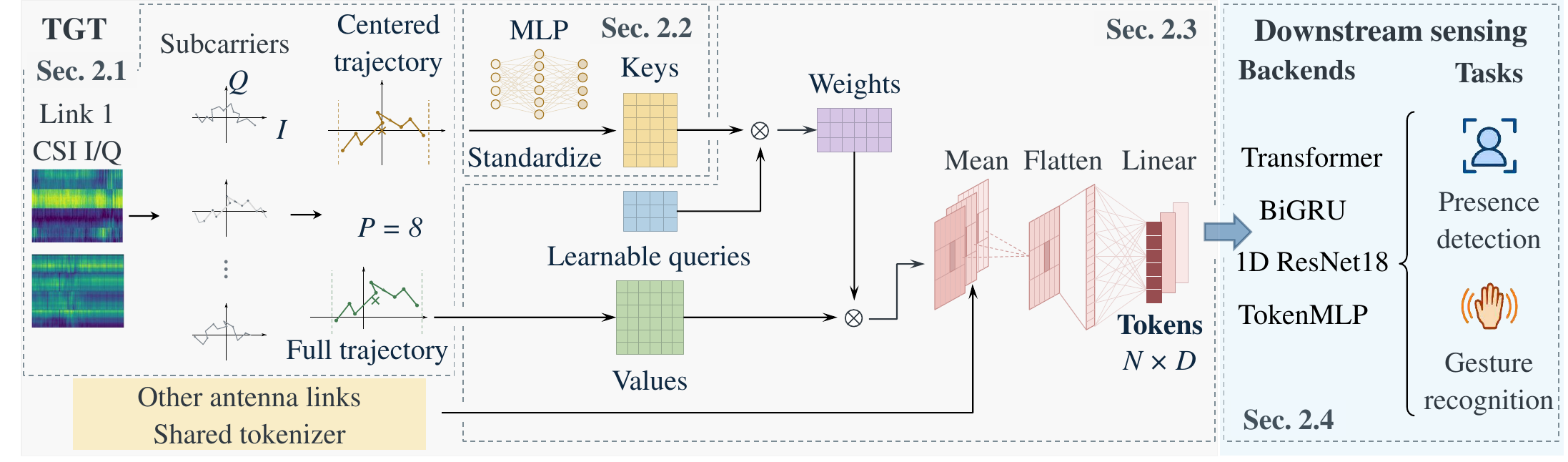}
\endgroup
\par\vspace{-2.5mm}
\caption{Overview of TGT, with dashed boxes corresponding to Sections~2.1--2.4.
One antenna link is shown using $P=8$ samples per patch; all links share the tokenizer.}
\label{fig:tgt_overview}
\end{figure*}

\section{Introduction}
\label{sec:intro}

Wi-Fi sensing uses channel state information (CSI) to capture fine-grained variations
in wireless propagation, enabling device-free recognition of human states and activities~\cite{SenseFi2023}.
CSI describes complex-valued channel responses across subcarriers and antenna links,
with successive measurements forming time series of amplitude and phase~\cite{SHARP2023}.
Human activity alters multipath propagation, inducing temporal variations in CSI~\cite{FarSense2019}
that deep neural networks learn to associate
with human states and activities~\cite{SenseFi2023}. However, models trained under
specific acquisition conditions may not generalize reliably to new environments,
users, or device placements~\cite{Widar2019,FewSense2024,WiFiJEPA2026,SHARP2023}. As CSI reflects both
the sensing target and propagation conditions, input representation is an important
consideration for cross-domain generalization.

Existing CSI-based sensing methods such as SHARP~\cite{SHARP2023} and
SLNet~\cite{SLNet2023} use motion-related representations derived from CSI,
including Doppler traces and spectrograms, as inputs to convolutional networks.
Other approaches instead encode CSI measurements directly.
CSI-JEPA~\cite{CSIJEPA2026} encodes amplitude-based time-frequency patches but
cannot distinguish CSI sequences that differ only in phase.
In contrast, WiFi-JEPA~\cite{WiFiJEPA2026} jointly embeds
CSI amplitude and denoised phase while preserving time and antenna-link structure.
However, its cross-environment generalization remains limited~\cite{WiFiJEPA2026}.
These approaches determine what information enters the model and how it is
organized, but leave open which information should guide subcarrier weighting
and which should be combined to form input representations. This motivates us
to study CSI tokenization as a representation-learning problem: encoding complex
observations into \textbf{continuous feature vectors} (i.e., tokens) to support
diverse sensing tasks and improve generalization across environments.

To guide the weighting and combination of CSI observations, we view each
subcarrier's short, ordered sequence of real and imaginary (I/Q) coordinates as a trajectory in the complex
plane. Our central insight is that centered trajectories can guide subcarrier
aggregation, while both local centers and centered trajectories are retained as
the content to be aggregated. Similar temporal variations can occur around different local centers,
which may vary across acquisition conditions. Using uncentered trajectories to
determine subcarrier aggregation weights may therefore make these weights sensitive
to shifts in local centers across environments.
Removing these centers entirely, however, may discard useful context or slowly varying information.
\textbf{Related time-series methods also address the trade-off between removing statistical
variation and preserving useful information}: RevIN removes and restores instance
statistics, while Non-stationary Transformers compensates for information lost
through stationarization~\cite{RevIN2022,Nonstationary2022}.

We propose Trajectory-Guided Tokenization (TGT), a complex CSI tokenizer trained jointly
with the downstream sensing model. TGT uses asymmetric cross-attention, with keys learned
from centered trajectories and values retaining both local complex means and trajectories.
With TokenMLP, TGT achieves mean accuracies of 92.83\% on the self-collected
dataset and 90.44\% on EHUNAM, supporting its applicability to cross-domain
presence detection across datasets. Evaluation on Widar extends this evidence
to a different sensing task, with TGT reaching 72.14\% mean accuracy for
cross-room gesture recognition with ResNet18.

\section{Trajectory-Guided CSI Tokenization}
\label{sec:tgt}

Figure~\ref{fig:tgt_overview} illustrates how TGT maps each temporal patch of
complex CSI spanning all subcarriers and antenna links to one continuous
token. TGT decomposes local trajectories into centers and centered
coordinates (Section~\ref{ssec:tgt_content}), constructs routing keys from
centered trajectories (Section~\ref{ssec:tgt_key}), and aggregates both local centers
and centered-trajectory coordinates through frequency routing, link pooling, and projection
(Section~\ref{ssec:tgt_routing}). The resulting tokens serve as inputs to
downstream sensing models (Section~\ref{ssec:tgt_downstream}).

\subsection{Complex Trajectory Decomposition}
\label{ssec:tgt_content}
\enlargethispage{-2.5mm}

Let $\mathbf X\in\mathbb R^{B\times T\times C\times F\times2}$ denote the
preprocessed CSI input (Section~\ref{sec:experiments}), expressed in real and
imaginary (I/Q) coordinates,
where $B$ is the batch size, and $T$, $C$, and $F$ denote the numbers of time
steps, antenna links, and subcarriers, respectively. We partition each CSI
window along the temporal axis into $N$
non-overlapping patches of $P$ consecutive time steps,
with $T=NP$. Specifically,
$\mathbf Z_f\in\mathbb R^{P\times2}$ contains the $P$ consecutive I/Q
observations of subcarrier $f$ on antenna link $\ell$ within temporal patch
$n$ of batch element $b$.

We decompose each local complex trajectory into its \textbf{center and centered
trajectory coordinates} using a fixed orthonormal Helmert matrix~\cite{Lancaster1965}:
\begin{equation}
\mathbf H_P=
\begin{bmatrix}
\mathbf1_P^\top/\sqrt P\\
\mathbf B_P
\end{bmatrix},
\label{eq:tgt_helmert_basis}
\end{equation}
where $\mathbf B_P\in\mathbb R^{(P-1)\times P}$ contains the contrast rows,
with $\mathbf B_P\mathbf1_P=\mathbf0$ and
$\mathbf H_P^\top\mathbf H_P=\mathbf I_P$.
The local complex mean of the patch is
\begin{equation}
\mu_f=\frac{1}{P}\sum_{p=0}^{P-1}
\left(I_{f,p}+\mathrm{i}Q_{f,p}\right),
\label{eq:tgt_patch_mean}
\end{equation}
where $(I_{f,p},Q_{f,p})$ is row $p+1$ of $\mathbf Z_f$. Applying the
transform to both I/Q columns gives
\begin{equation}
\mathbf Y_f=\mathbf H_P\mathbf Z_f=
\begin{bmatrix}
\sqrt P\,\operatorname{Re}\mu_f & \sqrt P\,\operatorname{Im}\mu_f\\
\mathbf d_f^{\mathrm{I}} & \mathbf d_f^{\mathrm{Q}}
\end{bmatrix},
\label{eq:tgt_helmert_transform}
\end{equation}
where the contrast vectors
$\mathbf d_f^{\mathrm{I}},\mathbf d_f^{\mathrm{Q}}\in\mathbb R^{P-1}$
encode the centered trajectory in an orthonormal basis. We concatenate the I/Q contrast
vectors to form $\mathbf r_f$, then combine $\mathbf r_f$ with the center
coordinates to obtain the center--trajectory representation $\mathbf c_f$:
\begin{gather}
\mathbf r_f=
\big[(\mathbf d_f^{\mathrm{I}})^\top,\;(\mathbf d_f^{\mathrm{Q}})^\top\big]^\top
\in\mathbb R^{2P-2},
\label{eq:tgt_trajectory_content}\\
\mathbf c_f=
\big[\sqrt P\,\operatorname{Re}\mu_f,\;
\sqrt P\,\operatorname{Im}\mu_f,\;\mathbf r_f^\top\big]^\top
\in\mathbb R^{2P}.
\label{eq:tgt_content}
\end{gather}
The inverse $\mathbf Z_f=\mathbf H_P^\top\mathbf Y_f$ recovers the ordered
I/Q patch, so $\mathbf c_f$ retains all information in $\mathbf Z_f$.

\subsection{Centered-Trajectory Keys}
\label{ssec:tgt_key}
\enlargethispage{-2.5mm}

Routing keys are constructed from the \textbf{centered trajectory coordinates}
$\mathbf r_f$. We standardize each coordinate using training-set statistics
with a variance floor $\epsilon$. A shared MLP maps the
standardized coordinates $\widetilde{\mathbf r}_f$ to trajectory features:
\begin{equation}
\label{eq:tgt_key_mlp}
\overline{\mathbf k}_f=
\operatorname{MLP}_k(\widetilde{\mathbf r}_f)
\in\mathbb R^{d_k}.
\end{equation}
The MLP has two affine layers, hidden width $d_h$, and an intermediate GELU
activation, with weights shared across temporal patches, antenna links, and
subcarriers. We then add a fixed sinusoidal encoding of the subcarrier index
to obtain the routing key:
\begin{equation}
\label{eq:tgt_key_position}
\mathbf k_f=\overline{\mathbf k}_f+\operatorname{PE}(f),
\qquad \operatorname{PE}(f)\in\mathbb R^{d_k}.
\end{equation}


\subsection{Asymmetric Frequency Routing}
\label{ssec:tgt_routing}

Asymmetric frequency routing (AFR) uses $K$ learnable queries
$\mathbf q_m\in\mathbb R^{d_k}$, $m=1,\ldots,K$, shared across CSI windows,
temporal patches, and antenna links. The \textbf{center--trajectory coordinates}
serve as Values, $\mathbf v_f=\mathbf c_f$, retaining the local centers
excluded from Keys. For each antenna link and
temporal patch of $P$ consecutive time steps, asymmetric
cross-attention~\cite{Vaswani2017} compresses the value vectors
$\mathbf v_f$ of all $F$ subcarriers into $K$ frequency slots:
\begin{gather}
\alpha_{m,f}=
\frac{\exp(\mathbf q_m^\top\mathbf k_f/\sqrt{d_k})}
{\sum_{f'=1}^{F}\exp(\mathbf q_m^\top\mathbf k_{f'}/\sqrt{d_k})},
\label{eq:tgt_frequency_routing}\\
\mathbf u_m=\sum_{f=1}^{F}\alpha_{m,f}\mathbf v_f
\in\mathbb R^{2P}.
\label{eq:tgt_slot_content}
\end{gather}
For each temporal patch, mean pooling combines corresponding frequency
slots across valid antenna links. The pooled slots are then flattened in
query order and mapped to a $D$-dimensional temporal token through a learned
linear layer.

\subsection{Downstream Learning}
\label{ssec:tgt_downstream}

The temporal tokens generated in Section~\ref{ssec:tgt_routing} serve as inputs
to the evaluated downstream architectures: Transformer, BiGRU, 1D ResNet18,
and TokenMLP (Section~\ref{sec:experiments}). We append a linear classification
head to each backend for end-to-end supervised training from scratch with
cross-entropy. The Helmert basis, frequency encodings, and calibration
statistics remain fixed.

\begin{figure*}[!t]
\centering
\includegraphics[width=\textwidth]{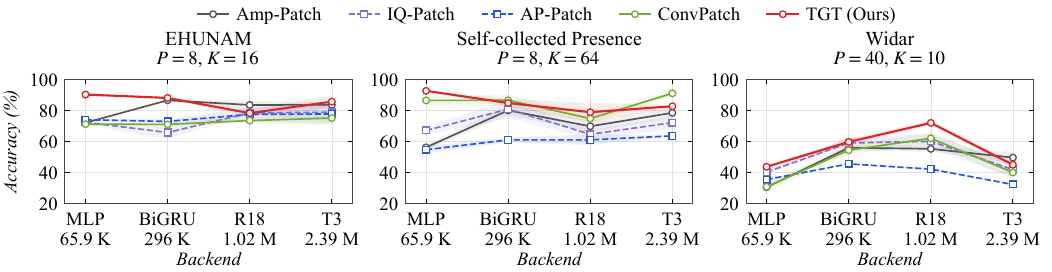}
\par\vspace{-8pt}
\caption{Accuracy comparison of TGT and CSI encoding baselines across three datasets.
Lines show means; shaded bands indicate $\pm$ one sample standard deviation.}
\label{fig:tokenizer_comparison_metrics}
\end{figure*}

\afterpage{%
\begin{table}[!t]
\vspace{-7pt}
\centering
\fontsize{9}{10.5}\selectfont
\caption{Tokenizer size (baseline $-$ TGT).}
\label{tab:model_parameters}
\vspace{3pt}
\setlength{\tabcolsep}{4pt}
\renewcommand{\arraystretch}{1.1}
\renewcommand{\tabularxcolumn}[1]{m{#1}}
\begin{tabularx}{\columnwidth}{@{}>{\centering\arraybackslash}m{68pt}|*{4}{>{\centering\arraybackslash}X}@{}}
\hline
Dataset & Amp & AP & IQ & Conv \\
\hline
Self-collected & $-0.8$k & $-0.8$k & $-0.8$k & $+2.3$k \\
\hline
EHUNAM & $-0.9$k & $-0.9$k & $-0.9$k & $+2.9$k \\
\hline
Widar & $-30.2$k & $-30.2$k & $-30.2$k & $-11.7$k \\
\hline
\end{tabularx}
\end{table}
}

\section{Experiments}
\label{sec:experiments}

\subsection{Datasets}
\label{ssec:datasets}

\textbf{Self-collected presence dataset.} We evaluate human presence detection
on a self-collected dataset comprising 157 GB of CSI recordings, acquired in
multiple collection campaigns across different cities for an industrial
presence-sensing application. The recordings capture human activities at varying
distances, including walking, standing up, body turning, and small arm or head
movements while seated. Environmental disturbances include moving curtains and
electric fans, recorded both with and without human presence. CSI is
collected in a monostatic sensing setup with three antenna links, 1,024
subcarriers over an 80 MHz channel, and a temporal sampling rate of 200 Hz.
We use room-disjoint training, validation, and test splits comprising 143,
68, and 93 recordings from seven, three, and four rooms, respectively.

\textbf{EHUNAM.} We further evaluate presence detection using the 80~MHz MC1
recordings from EHUNAM~\cite{EHUNAM2025}.
The processed CSI retains 241 subcarriers and is resampled to 100~Hz, with
receiver-specific recordings treated as separate inputs. Training uses recordings from
the basement, laboratory, and storage room; validation and testing use the
classroom and office, respectively. These splits contain 500, 187, and 166
receiver-specific CSI recordings, respectively.

\textbf{Widar3.0.} To assess applicability beyond presence detection,
we evaluate TGT on Widar3.0 for cross-room gesture
recognition. CSI is collected using six receivers at 1,000 Hz, with
30 subcarriers per antenna link~\cite{Widar2019}. Rooms 1 and 3 provide
training and validation data, while Room 2 is reserved for testing. All
receiver views of the same gesture instance remain in the same split.
This six-class evaluation uses 4,169 training,
1,038 validation, and 737 test recordings.

\fixedsubsection{Experimental Setup}
\label{ssec:experimental_setup}

For each self-collected CSI window, we subtract the complex mean over time
separately for each antenna link and subcarrier, then apply shared-scale I/Q
normalization and radial amplitude clipping.
CSI processing for EHUNAM and Widar3.0 is based on the procedures described
in their respective studies~\cite{EHUNAM2025,Widar2019}.

Window length and stride are both 256 time steps for the self-collected
dataset and 200 for EHUNAM. For each dataset, all frontends use the same
temporal patch size $P$ and produce one 256-dimensional token per
patch, with $P$ and $K$ specified in Figure~\ref{fig:tokenizer_comparison_metrics}.
TGT uses a two-layer Key MLP with dimensions $14\rightarrow32\rightarrow14$
for both presence datasets and $78\rightarrow192\rightarrow78$
for Widar. Models are jointly trained
from scratch in PyTorch for 30 epochs using
AdamW~\cite{Loshchilov2019} with a batch size of 64. We repeat experiments with multiple random seeds and
report the mean and sample standard deviation of test accuracy
over the runs included in each comparison.

\begingroup
\setlength{\parskip}{0pt}
\fixedsubsection{Tokenizer Comparison}
\label{ssec:tokenizer_comparison}

We compare TGT with four \textbf{frontend baselines}.
Amp-Patch uses amplitude-only inputs, consistent with
CSI-JEPA~\cite{CSIJEPA2026} and AM-FM~\cite{AMFM2026};
IQ-Patch uses real and imaginary components;
AP-Patch uses amplitude and phase, as in WiFi-JEPA~\cite{WiFiJEPA2026}.
The three patch baselines preserve temporal order and use cross-attention across
subcarriers, with Keys and Values derived from the same input coordinates.
ConvPatch applies ViT-style linear patch embedding to time--frequency I/Q
patches through a shared 2D convolution~\cite{WiFo2024,Dosovitskiy2020}.

For the \textbf{backends}, we use lightweight architectures so that
classification performance more directly reflects tokenizer effectiveness.
TokenMLP processes tokens independently, whereas
Transformer-3, BiGRU, and 1D ResNet18 model temporal dependencies across
patches. With dataset-specific $P$, $K$, and Key MLP dimensions, TGT contains
264.2k, 67.0k, and 236.1k parameters for the self-collected, EHUNAM, and
Widar datasets, respectively. Its size is comparable to the baseline
frontends on both presence datasets (Table~\ref{tab:model_parameters}).

On the self-collected dataset, TGT achieves the highest mean
accuracy among the evaluated frontend--backend combinations in
Figure~\ref{fig:tokenizer_comparison_metrics}, reaching
\textbf{92.83\%} with TokenMLP, compared with 86.65\% for ConvPatch with
the same backend. TGT also achieves the highest mean accuracy with ResNet18,
while ConvPatch performs better with BiGRU and Transformer-3, reflecting
differences in tokenizer rankings across backends.
Retaining phase information alone does not guarantee better cross-domain
performance: IQ-Patch underperforms Amp-Patch with Transformer-3 and ResNet18,
reaching 64.82\% versus 69.95\% accuracy with the latter.
AP-Patch records the lowest mean accuracy with all four
backends, motivating further examination of phase quality and
representation-specific encoding~\cite{WiFiJEPA2026,SHARP2023}.

On EHUNAM, TGT with TokenMLP achieves the highest mean accuracy
(\textbf{90.44\%}), compared with 74.23\% for AP-Patch, the strongest
baseline with the same backend. On Widar, TGT with ResNet18 also achieves
the highest mean accuracy (\textbf{72.14\%}), exceeding ConvPatch (62.20\%)
with the same backend. These results extend the presence-detection
findings to a second dataset and support TGT's applicability to cross-room
gesture recognition.
\par\endgroup

\begingroup
\setlength{\parskip}{0pt}
\fixedsubsection{Ablation Study}
\label{ssec:ablation_study}

\begin{table}[!t]
\vspace{-7pt}
\centering
\fontsize{9}{10.5}\selectfont
\caption{TGT ablation accuracy (\%), mean $\pm$ sample standard deviation.}
\label{tab:routing_ablation}
\vspace{6pt}
\setlength{\tabcolsep}{4.5pt}
\renewcommand{\arraystretch}{1.5}
\renewcommand{\tabularxcolumn}[1]{m{#1}}
\begin{tabularx}{\columnwidth}{@{}>{\centering\arraybackslash}m{55pt}|*{3}{>{\centering\arraybackslash}X}@{}}
\hline
Variant & Presence & EHUNAM & Widar \\
\hline
w/o Helmert & $64.93\pm3.37$ & $70.69\pm1.64$ & $60.02\pm2.01$ \\
\hline
w/~Key~Center & $90.86\pm0.46$ & $87.80\pm5.93$ & $70.33\pm2.31$ \\
\hline
w/o~V~Center & $89.62\pm1.46$ & $87.53\pm1.15$ & $70.96\pm1.08$ \\
\hline
\textbf{TGT} & $\mathbf{92.83\pm0.35}$ & $\mathbf{90.44\pm1.02}$ & $\mathbf{72.14\pm1.41}$ \\
\hline
\end{tabularx}
\end{table}

We compare full TGT with three variants to examine the coordinate
representation and the roles of local centers, using TokenMLP for both
presence datasets and ResNet18 for Widar (Table~\ref{tab:routing_ablation}).
The w/o~Helmert variant uses ordered I/Q as both the Key MLP input and
Values, removing the transform and allowing local centers to influence
Key learning. The w/~Key~Center variant retains the Helmert transform but
includes both center and trajectory coordinates in the Key MLP input,
leaving Values unchanged. The w/o~V~Center variant leaves Key construction
unchanged and sets the two center coordinates in Values to zero, without
changing the Value dimension.

The w/~Key~Center variant achieves higher
mean accuracy than w/o~Helmert across all three
datasets, supporting the use of an explicit center--trajectory representation.
TGT further exceeds w/~Key~Center by 1.97, 2.64, and 1.81 percentage points
in mean accuracy on the self-collected dataset, EHUNAM, and Widar,
respectively, suggesting that excluding local centers from routing
can improve performance. In contrast, w/o~V~Center has lower mean accuracy on all
three datasets, supporting the retention of local centers in Values to
preserve useful predictive information.
\par\endgroup

\section{Conclusion}
\label{sec:conclusion}

We presented TGT, a learnable frontend that constructs continuous tokens
from complex CSI. An orthonormal Helmert transform provides an explicit
center--trajectory representation for asymmetric cross-attention, with Keys
learned from centered-trajectory coordinates and Values retaining both the
center and trajectory components. Trained jointly from scratch, TGT achieves
the highest mean accuracy among the evaluated frontend--backend combinations
for cross-domain presence detection on both presence datasets with TokenMLP
and for cross-room gesture recognition on Widar with ResNet18. The ablations
further support these design choices. These findings suggest that information
retained for prediction need not also guide aggregation.

%

\clearpage
\balance
\bibliographystyle{IEEEbib}
\bibliography{strings,refs}

\end{document}